\documentclass{article}
\usepackage{booktabs} 
\usepackage{graphicx}
\usepackage{PRIMEarxiv}
\usepackage{booktabs} 
\usepackage[utf8]{inputenc} 
\usepackage[T1]{fontenc}    
\usepackage{hyperref}       
\usepackage{url}            
\usepackage{booktabs}       
\usepackage{amsfonts}       
\usepackage{nicefrac}       
\usepackage{microtype}      
\usepackage{lipsum}
\usepackage{fancyhdr}       
\usepackage{graphicx}       
\graphicspath{{media/}}     

\title{MSTIM: A MindSpore-Based Model for Traffic Flow Prediction
\thanks{\textit{\underline{}}
\textbf{These authors contributed equally to this work.}} 
}

\author{
  Weiqi Qin\\
  Institute of Software \\
  Chinese Academy of Sciences   \\
  100190,China \\
  \texttt{weiqi@isrc.iscas.ac.cn} \\
   \And
  Yuxin Liu \\
  School of Information Technology \\
  Guangxi Police College   \\
  530028,China \\
  \texttt{liuyuxin@gxjcxy.edu.cn}
   \And
  Dongze Wu \\
  Institute of Software \\
  Chinese Academy of Sciences   \\
  100190,China \\
  \texttt{dongze@isrc.iscas.ac.cn} \\
  \And
  Zhenkai Qin \\
  School of Information Technology \\
  Guangxi Police College   \\
  530028,China \\
  \texttt{qinzhenkai@gxjcxy.edu.cn} \\
  \And
  Qining Luo \\
  Institute of Software \\
  Chinese Academy of Sciences   \\
  100190,China \\
  \texttt{qining@gxjcxy.edu.cn} \\
}

\begin{document}
\maketitle

\begin{abstract}
Aiming at the problems of low accuracy and large error fluctuation of traditional traffic flow predictionmodels when dealing with multi-scale temporal features and dynamic change patterns. this paperproposes a multi-scale time series information modelling model MSTIM based on the Mindspore
framework, which integrates long and short-term memory networks (LSTMs), convolutional neural
networks (CNN), and the attention mechanism to improve the modelling accuracy and stability. The
Metropolitan Interstate Traffic Volume (MITV) dataset was used for the experiments and compared
and analysed with typical LSTM-attention models, CNN-attention models and LSTM-CNN models.
The experimental results show that the MSTIM model achieves better results in the metrics of Mean
Absolute Error (MAE), Mean Square Error (MSE), and Root Mean Square Error (RMSE), which
significantly improves the accuracy and stability of the traffic volume prediction.
\end{abstract}


\section{Introduction}
With the continuous development of artificial intelligence technology, advanced methods such as deep learning have been increasingly applied in the field of transport, promoting the rapid evolution of intelligent transport systems. Traffic flow prediction, as one of the important research directions, plays a key role in improving the operational efficiency of road networks, alleviating traffic congestion, and optimising the overall traffic management, which are directly related to the quality of life and the economic operation level of the city \cite{1} . Highly accurate prediction results help to achieve intelligent regulation of traffic signals, dynamic optimisation of travel paths, and time-sharing charges based on congestion conditions, thus effectively improving road capacity and reducing travel delays \cite{2} . At present, while traffic flow has strong spatial and temporal characteristics, traffic flow prediction should consider both the long-term correlation of traffic flow in the time and space dimensions and the potential correlation that exists between traffic flows in different locations \cite{3} .However, existing traffic flow prediction methods generally suffer from low prediction accuracy and large error fluctuations.

In order to solve the above problems, this paper introduces a multi-scale temporal information modelling method, MSTIM, based on the MindSpore framework, which integrates convolutional neural network (CNN), long and short-term memory network (LSTM), and Attention to improve the ability of modelling complex changes in traffic flow. The model incorporates Convolutional Neural Network (CNN), Long Short Term Memory Network (LSTM) and Attention to improve the modelling of complex traffic flow changes. Firstly, the CNN module is used to extract local spatial features in the traffic flow data to effectively capture short-term changes; secondly, the LSTM network models long-term dependencies in the time series to enhance the model's understanding of trending changes; and finally, the Attention mechanism is introduced to dynamically weight the key time points in the time series to enhance the response capability to sudden events. The experimental results show that the proposed MSTIM model outperforms the traditional methods and mainstream deep learning models on the Metro Interstate Traffic Volume dataset, and achieves better performance in the indicators of MAE, RMSE and MAPE, which verifies the validity and practicability of the proposed method in the traffic volume prediction task.

\section{Related Work}
Traffic flow forecasting is the process of speculating future traffic conditions at a specific time and location by analysing historical traffic data \cite{4,5}, aiming to simulate the complex spatio-temporal dependencies inherent in the system, and thus accurately predicting changes in traffic states \cite{6,7}. In order to improve the accuracy and reliability of traffic flow prediction, scholars have conducted extensive research on it, and the studies are mainly divided into three categories. 
Prediction models based on traditional statistical methods, such as autoregressive integral sliding average (ARIMA), historical averaging (HA), etc.; models based on machine learning algorithms, such as support vector machines (SVMs), random forests (RFs), K nearest neighbours (KNN), etc.; and prediction models based on deep learning, such as long- and short-term memory networks (LSTMs), recurrent neural networks (GNNs), and Convolutional Neural Network (CNN), etc. 

Traditional statistical methods mainly build mathematical models for prediction by analysing the temporal characteristics of historical traffic data. For example, Hamed et al \cite{8} accurately predicted the future traffic flow of urban arterial roads using the ARIMA model, which describes the process of traffic flow through mathematical modelling.Smith et al \cite{9} used the HA method for predicting the urban traffic flow in their study, which was particularly applied to the urban traffic control system and the European Travel Information System (ETIS). The method predicts future traffic flow by weighted average of historical traffic data. Although the HA method is faster, it relies to a certain extent on the cyclical changes of traffic flow and fails to fully consider the temporal variability and potential instability of traffic flow. Therefore, in the complex and changing traffic environment, the prediction effect of HA method is usually unsatisfactory, and it is difficult to meet the practical application requirements.

With the improvement of computing power, machine learning algorithms have been widely used in traffic flow prediction. For example, Zhang et al \cite{10} proposed a KNN-based model to improve the accuracy of short-term traffic flow prediction, which constructed a short-term traffic flow prediction system for urban expressways from the historical database, search mechanism and algorithmic parameters, and the experimental results showed that the prediction accuracy of the method was more than 90\%. Similarly, SVR is a commonly used prediction method, for example, Ahn et al\cite{11} further advanced the field by integrating Bayesian classifiers with SVR for real-time traffic flow prediction. Their approach uses 3D Markov random fields to model the relationship between traffic flow and road conditions, demonstrating the potential of combining different statistical methods to enhance prediction performance.Ge et al \cite{12} proposed a multiresolution SVR prediction model based on the concept of local modelling, which can effectively fit the changes of local details, thus improving the prediction accuracy.

In recent years, deep learning techniques have made significant progress in the field of traffic flow prediction, especially represented by the application of neural network models such as CNN, LSTM, and RNN. The wide application of these models has triggered researchers to explore their integration in depth, with a view to improving the accuracy and efficiency of traffic prediction in different scenarios. For example, Wang et al \cite{13} introduced a novel spatial temporal graphical neural network (ST-GNN) designed to capture the complex spatial and temporal patterns inherent in traffic flow data. The model emphasises the importance of understanding the relationships between different locations and the temporal dependence of traffic for accurate prediction.Yuan et al \cite{14} developed a personalised traffic conflict prediction model using LSTM. This approach incorporates individual driving patterns and improves classification accuracy and generalisation, thus helping to develop targeted interventions to mitigate risky driving behaviours.Balasubramani et al \cite{15} focused on predicting bus patronage using a bidirectional LSTM fusion model that processes past and future features through a bidirectional LSTM layer. The prediction accuracy is improved compared to traditional prediction methods.Furthermore, Ibrahim et al \cite{16}  proposed a multilayer CNN-GRUSKIP model that combines CNN and GRU-SKIP mechanisms for spatio-temporal traffic flow prediction. The model extracts spatial features through deep CNN layers and handles long time dependencies using the GRU-SKIP mechanism, and the experimental results show that it outperforms the conventional models on several real datasets. However, although existing models have achieved good results, there are still some challenges, especially in capturing multi-scale, long-term dependencies, and complex spatio-temporal interactions in traffic flow data.
\section{Model Description}

\begin{figure}[htbp]
  \centering
\includegraphics[width=1\linewidth]{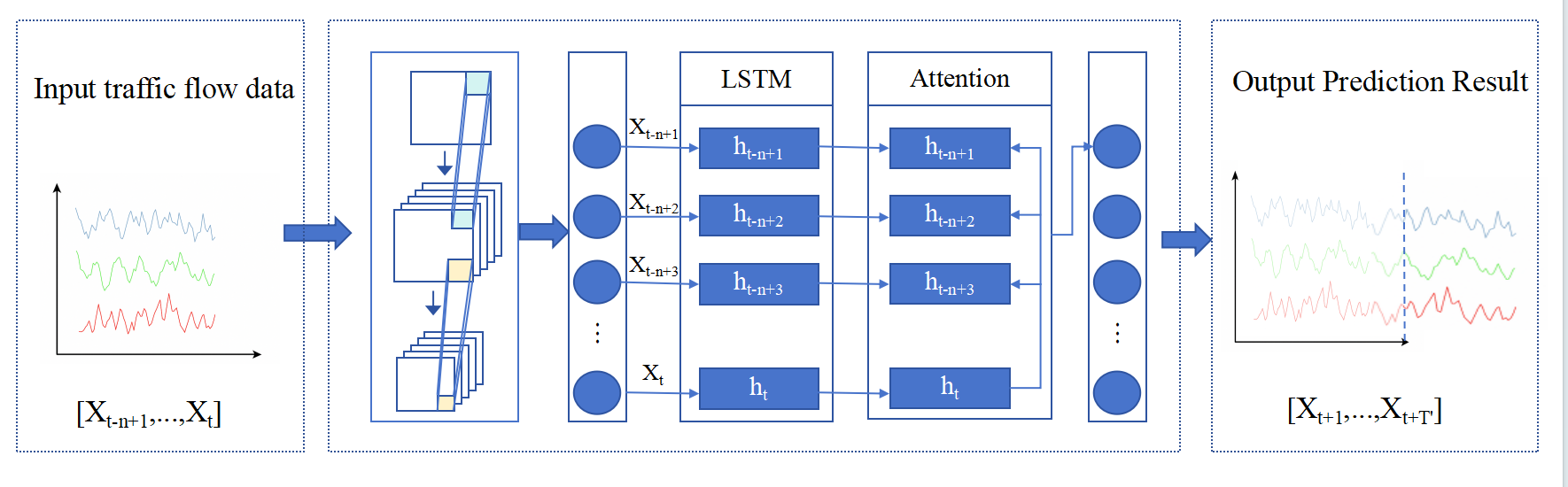}
  \caption{Structural diagram of the MSTIM traffic flow prediction model, which integrates convolutional neural network (CNN), long short-term memory network (LSTM) and attention mechanism to realize accurate prediction of future traffic flow.}
  \label{fig:model architecture}
\end{figure}

This study proposes a multi-scale temporal information model (MSTIM) based on the MindSpore framework, which integrates convolutional neural networks (CNN), long short-term memory (LSTM) networks, and an attention mechanism to enhance the accuracy and robustness of traffic flow prediction. The specific structure of the model is shown in Fig\ref{fig:model architecture}.The model takes the historical traffic flow sequence 
\([X_{t-n+1}, \dots, X_t]\) 
as input, and first utilizes a multi-scale CNN module to extract local spatial features, capturing variations across different temporal scales. These features are then fed into an LSTM network to model long-term temporal dependencies. To further improve performance, an attention mechanism is introduced to assign dynamic weights to critical time steps, enabling the model to focus on informative patterns such as sudden traffic fluctuations. Finally, the model predicts traffic flow for future time steps 
\([X_{t+1}, \dots, X_{t+T}]\), 
thereby achieving precise forecasting of traffic trends.
\subsection{LSTM}
\begin{figure}[htbp]
  \centering
  \includegraphics[width=0.7\linewidth]{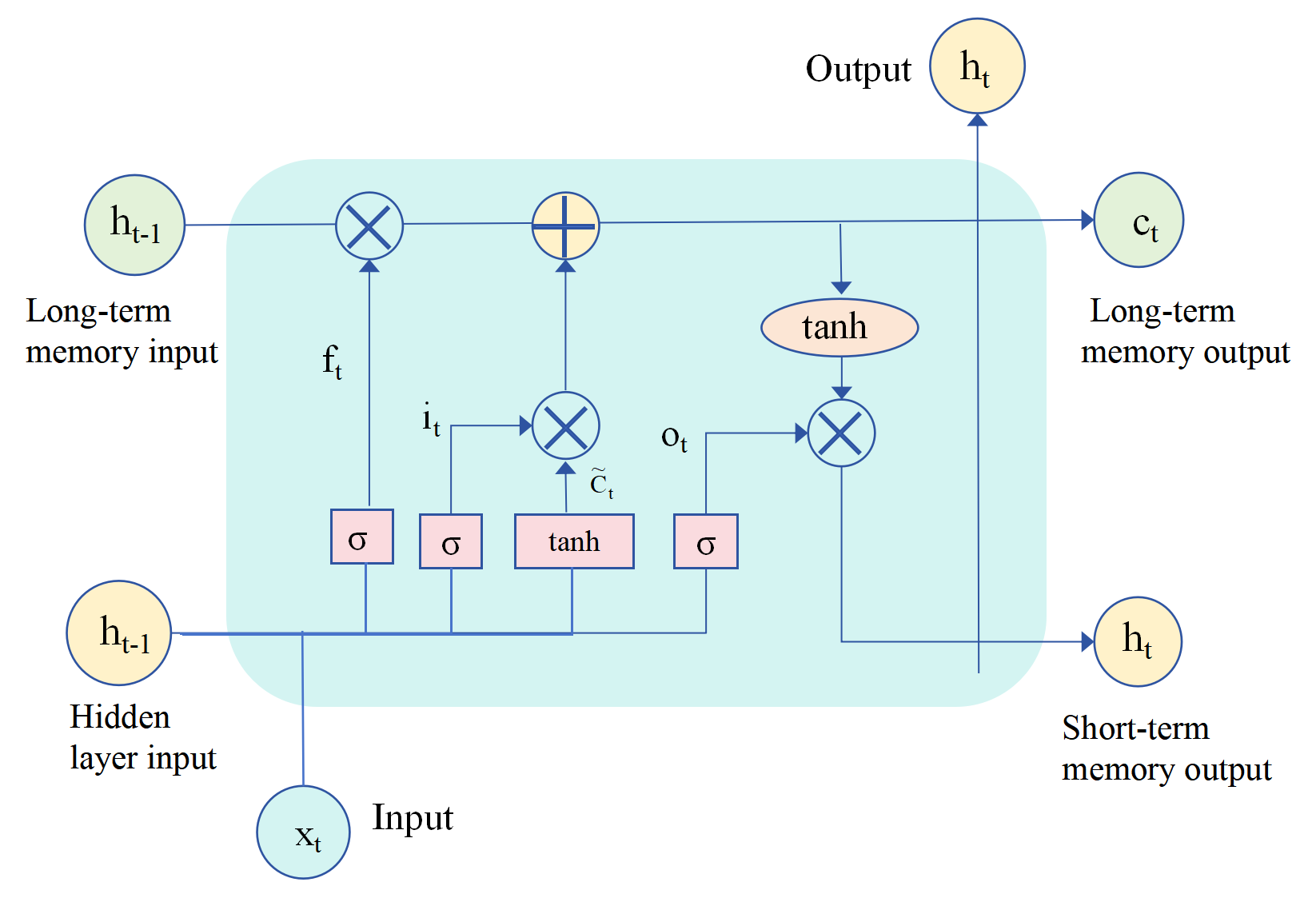}
  \caption{Structure of an LSTM cell. The diagram illustrates the internal flow of information, including the forget gate (\(f_t\)), input gate (\(i_t\)), candidate memory cell (\(\tilde{C}_t\)), output gate (\(o_t\)), and the updates of long-term memory (\(C_t\)) and hidden state (\(h_t\)).}
  \label{fig:LSTM model architecture}
\end{figure}

Long Short-Term Memory (LSTM) is a variant of Recurrent Neural Networks (RNN) designed to address the problem of gradient vanishing encountered by traditional RNNs when dealing with long-term dependencies.The specific model architecture is illustrated in Fig\ref{fig:LSTM model architecture}.LSTM controls the flow of information through the introduction of input gates, forgetting gates, and output gates, which can efficiently and selectively ‘remember’ or ‘forget’ information, thus enabling the capture of long-term dependencies in time series \cite{17}. In the traffic flow prediction task, LSTM employs a feedback mechanism whereby the output of the previous time step is used as the input of the current time step, thus propagating past information to future states. This feedback mechanism preserves temporal correlation and is therefore suitable for capturing the temporal evolution of traffic parameters \cite{18} The specific process is as follows:

\textbf{Input Gate:} The input gate, activated by a sigmoid function, controls the extent to which the current input and candidate memory \( \tilde{C}_t \) update the cell state.The formula is:

\begin{equation}
i_t = \sigma(W_i \cdot [h_{t-1}, x_t] + b_i)
\end{equation}

Where \( i_t \) is the output of the input gate, \( W_i \) is the weight matrix of the input gate, \( b_i \) is the bias term, \( h_{t-1} \) is the hidden state from the previous time step, and \( x_t \) is the input at the current time step.

\textbf{Forget Gate:} The forget gate, activated by a sigmoid function, controls how much of the previous memory is retained. The formula is:

\begin{equation}
f_t = \sigma(W_f \cdot [h_{t-1}, x_t] + b_f)
\end{equation}

Where \( f_t \) is the output of the forget gate, \( W_f \) is the weight matrix of the forget gate, and \( b_f \) is the bias term.

\textbf{Memory Cell:} The memory cell, as the core of LSTM, maintains state information over time. Its update is jointly controlled by the forget gate, which retains past information, and the input gate, which incorporates new input.
 The updated memory cell formula is:

\begin{equation}
C_t = f_t \cdot C_{t-1} + i_t \cdot \tilde{C}_t
\end{equation}

Where \( C_t \) is the current memory cell, \( C_{t-1} \) is the memory cell from the previous time step, and \( \tilde{C}_t \) is the candidate memory cell at the current time step.

\textbf{Output Gate:} The output gate, activated by a sigmoid function, controls how much of the current memory state contributes to the hidden state \( h_t \). The memory state is transformed by a \textit{tanh} function and modulated by the gate to produce the final output. The formula is:

\begin{equation}
h_t = o_t \cdot \tanh(C_t)
\end{equation}

Where \( h_t \) is the current hidden state, \( o_t \) is the output of the output gate, and \( C_t \) is the current memory cell.

\subsection{CNN}
\begin{figure}[htbp]
  \centering
  \includegraphics[width=0.9\linewidth]{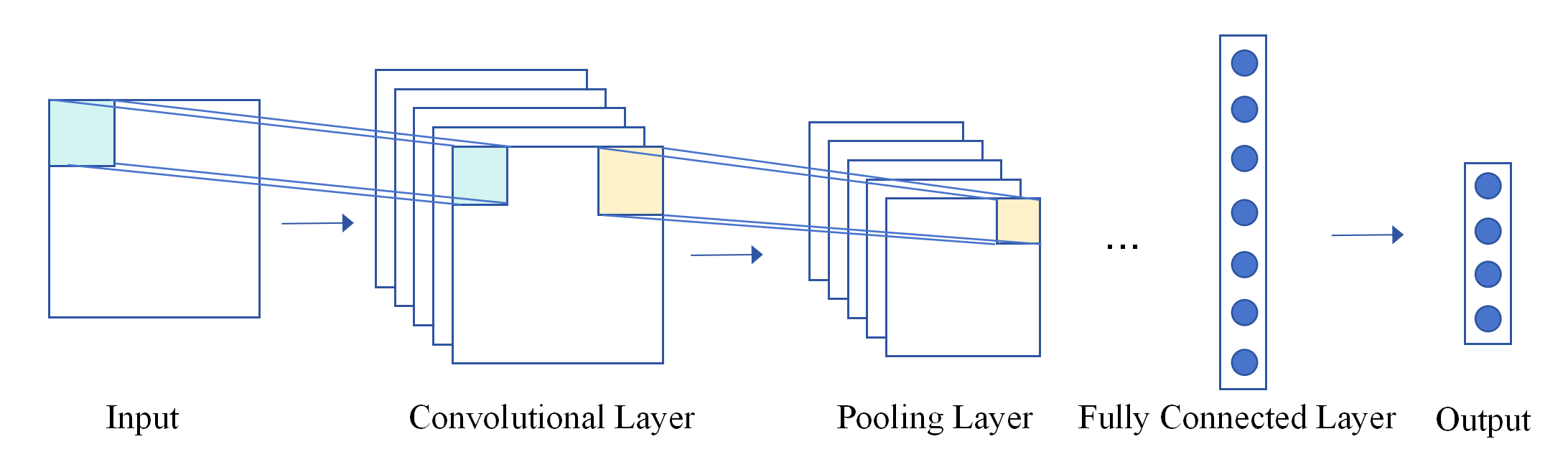}
  \caption{A schematic diagram of a convolutional neural network (CNN), illustrating the typical flow from input through convolutional and pooling layers to fully connected layers and final output.}
  \label{fig:CNN model architecture}
\end{figure}

Convolutional Neural Networks (CNNs), as a type of feedforward neural network, are widely applied in various domains such as image recognition, natural language processing, and time series analysis due to their relatively low number of parameters and high computational efficiency. In the context of traffic flow prediction, CNNs are particularly effective in capturing spatial dependencies within traffic data, thereby enhancing the model's ability to learn correlations across different regions of a road network. In this study, a CNN-based model is employed to extract spatial features of traffic flow patterns, and the overall architecture is illustrated in Fig\ref{fig:CNN model architecture}.

\textbf{Convolutional Layer:} The convolutional layer is the core of CNNs, responsible for extracting local features from the input using convolutional kernels. The operation is defined as:

\begin{equation}
Y(i, j) = (X * W)(i, j) = \sum_m \sum_n X(i + m, j + n) W(m, n)
\end{equation}
where \( X \) denotes the input feature map, \( W \) is the convolutional kernel, and \( Y \) is the output feature map.

\textbf{Activation Function:} Activation functions introduce non-linearity, enhancing the model's expressive power. A commonly used function is ReLU, defined as:

\begin{equation}
f(x) = \max(0, x)
\end{equation}
which effectively mitigates the vanishing gradient problem.

\textbf{Pooling Layer:} Pooling reduces the spatial dimensions of feature maps while retaining critical information. For example, a \(2 \times 2\) max pooling operation is:

\begin{equation}
Y(i, j) = \max(X(i, j), X(i+1, j), X(i, j+1), X(i+1, j+1))
\end{equation}

\textbf{Fully Connected Layer:} This layer flattens the extracted features and performs linear transformation for prediction tasks:

\begin{equation}
y = Wx + b
\end{equation}
where \( W \) is the weight matrix, \( x \) is the input vector, \( b \) is the bias, and \( y \) is the output.

\subsection{Attention}
Attention Mechanism plays a crucial role in time-series data modelling, especially in the task of traffic flow prediction. While traditional models usually rely on fixed weights or window sizes for historical data, Attention Mechanism greatly enhances the flexibility and sensitivity of the model to time series changes by dynamically ‘focusing’ on the most relevant time steps. This mechanism enables the model to effectively identify critical moments of traffic changes, such as sudden congestion during peak hours, or abnormal fluctuations in traffic under special weather conditions. The self-attention mechanism dynamically adjusts the focus of the model by calculating the similarity of each time point in the input sequence with other time points and assigning different weights to each time point, so that it can focus on the most informative parts and thus improve the prediction accuracy. The specific calculation formula is as follows:

\begin{equation}
\mathrm{Attention}(Q, K, V) = \mathrm{softmax}\left( \frac{QK^{\top}}{\sqrt{d_k}} \right) V
\end{equation}

where \( Q \) is the query vector, \( K \) is the key vector, \( V \) is the value vector, and \( d_k \) is the dimension of the keys.

\section{Experimental Results and Analysis}
\subsection{Dataset}
The Metro Interstate Traffic Volume dataset\cite{19} is a publicly available dataset collected by the Minnesota Department of Transportation for traffic flow forecasting. The dataset contains 48,205 records, which not only covers the traffic volume data at each time point, but also collects rich meteorological information, including temperature, rainfall, snowfall, cloud coverage, major weather conditions and their detailed descriptions, etc., with precise time stamps, which is widely used in the research on the relationship between traffic volume prediction and meteorological factors and traffic changes. It is widely used in traffic flow prediction and the study of the relationship between meteorological factors and traffic changes.
\subsection{Experimental Environment}
The experiment was conducted using the MindSpore framework on the Ubuntu 22.04.4 LTS operating system. The hardware platform was configured with an Intel® Xeon® Platinum 8358 processor (2.60 GHz), 1 TB of memory, and an NVIDIA Tesla V100-SXM2-32GB GPU, with CUDA version 12.4. To ensure the stability and efficiency of the model training process, the experimental parameters were carefully configured, as shown in Table\ref{table1}.
\begin{table}[htbp]
\centering
\caption{Main Training Parameters of the Model}
\label{table1}
\begin{tabular}{p{4cm} p{2cm} p{8cm}}
\toprule
\textbf{Parameter Name} & \textbf{Value} & \textbf{Description} \\
\midrule
Epochs & 10 & Total number of passes through the entire training dataset during model training. \\
Learning Rate & 0.001 & The step size used to update model parameters in the optimization algorithm. \\
Batch Size & 32 & Number of samples used in one iteration of model parameter update. \\
Kernel Sizes & [3, 5, 7] & Sizes of convolution kernels used to capture multi-scale spatial features. \\
\bottomrule
\end{tabular}
\end{table}
\subsection{Evaluation Metrics}
Three commonly used prediction evaluation metrics are used in this paper: mean absolute error (MAE), root mean square error (RMSE) and mean absolute percentage error (MAPE). Among them, MAE measures the average of the absolute errors between predicted and actual values, and a smaller MAE value indicates a higher prediction accuracy of the model, so it is an intuitive and widely used error metric. MSE, on the other hand, calculates the average of the squares of the errors between predicted and actual values, and it gives a higher penalty for larger errors, and a larger MSE value usually indicates that there is a larger prediction bias in the model. RMSE is the square root of MSE, and its unit is the same as that of the original data, which can reflect the size of prediction error more intuitively.

\begin{equation}
\mathrm{MAE} = \frac{1}{n} \sum_{i=1}^{n} \left| y_i - \hat{y}_i \right|
\end{equation}

\begin{equation}
\mathrm{RMSE} = \sqrt{ \frac{1}{n} \sum_{i=1}^{n} (y_i - \hat{y}_i)^2 }
\end{equation}

\begin{equation}
\mathrm{MSE} = \frac{1}{n} \sum_{i=1}^{n} (y_i - \hat{y}_i)^2
\end{equation}

Where \( y_i \) is the actual value, \( \hat{y}_i \) is the predicted value, and \( n \) is the number of data points. These metrics provide multi-dimensional evaluation criteria for the predictive accuracy of the model, helping us to comprehensively understand its performance. In particular, MAE, RMSE, and MSE each have their own unique applicability in reflecting different aspects of prediction error.

\subsection{Experimental Results and Analysis}
In order to verify the performance advantages of the MSTIM models proposed in this paper in the traffic flow prediction task, three deep learning models, LSTM-Attention, CNN-Attention and LSTM-CNN, are selected as comparative benchmarks and trained and evaluated on the same dataset and experimental conditions. Table\ref{table2} shows the prediction results of each model on the test set, including three common metrics: 
\begin{table}[htbp]
\centering
\caption{Model Performance Comparison}
\label{table2} 
\begin{tabular}{p{4cm} p{2.5cm} p{2.5cm} p{2.5cm}}
\toprule
\textbf{Model} & \textbf{MAE} & \textbf{MSE} & \textbf{RMSE} \\
\midrule
LSTM-Attention & 0.2570 & 0.1128 & 0.3369 \\
CNN-Attention  & 0.2358 & 0.1128 & 0.3399 \\
LSTM-CNN       & 0.2271 & 0.1101 & 0.3465 \\
MSTIM (ours)   & \textbf{0.2120} & \textbf{0.1048} & \textbf{0.3237} \\
\bottomrule
\end{tabular}
\end{table}

The results demonstrate that the MSTIM model achieves the best performance across all evaluation metrics. Its Mean Absolute Error (MAE) is 0.2120, which is significantly lower than that of LSTM-Attention (0.2570), CNN-Attention (0.2358), and LSTM-CNN (0.2271), representing a 17.5\% reduction compared to the worst-performing LSTM-Attention model. This indicates that MSTIM can more effectively reduce overall prediction bias and exhibits superior accuracy.In terms of Mean Squared Error (MSE), MSTIM records a value of 0.1048, showing varying degrees of reduction compared to other models, which suggests that it performs better in maintaining prediction stability and suppressing the amplification of errors caused by outliers. Notably, compared with LSTM-CNN (MSE = 0.1101), the error is reduced by approximately 4.8\%.Regarding Root Mean Squared Error (RMSE), MSTIM achieves the lowest value of 0.3237, outperforming LSTM-Attention (0.3369), CNN-Attention (0.3399), and LSTM-CNN (0.3465). This result suggests that MSTIM yields more consistent prediction outputs and demonstrates stronger robustness in handling complex traffic flow variations.

From the analysis of model structure level, the LSTM-Attention model only has the ability to model a single time dimension, although it strengthens the perception of key temporal nodes with the help of the attention mechanism, it lacks the effective modeling of spatial local features; the CNN-Attention model is able to capture spatial feature changes, but its ability to deal with long-term dependency is limited, which results in limited prediction stability; the LSTM-CNN model achieves joint modeling of temporal and spatial features, but its high RMSE value shows that the model still has insufficient error suppression ability when facing highly volatile traffic data. Although the LSTM-CNN model realizes the joint modeling of temporal and spatial features, its high RMSE value indicates that the model still lacks the error suppression ability when facing highly fluctuating traffic data. In contrast, the MSTIM model achieves more expressive deep modeling by introducing the LSTM network for long-time dependent modeling, integrating CNN to extract local spatial patterns, and using the attention mechanism to strengthen the time-step selection, while adopting a multi-scale temporal information processing strategy in the structural design to effectively integrate short-term fluctuation and long-term trend information. The experimental results show that MSTIM has significant advantages in prediction accuracy and error control, which verifies the effectiveness and practicality of the model in traffic flow prediction.

\section{Conclusion}
Aiming at the problems of low accuracy and large error fluctuation of traditional traffic flow prediction models when dealing with multi-scale temporal features and dynamic change patterns, this paper proposes a multi-scale temporal information modeling model, MSTIM, which integrates the long and short-term memory network (LSTM), convolutional neural network (CNN), and the attention mechanism, to achieve a good balance between prediction accuracy and error stability, through the introduction of the LSTM network for modeling the longterm dependency relationship between traffic flow data. The model models the long-term dependence of traffic flow data by introducing LSTM network, captures the local change features under different time granularity by using multi-scale CNN, and dynamically selects the key time step by combining with the attention mechanism, so as to achieve a good balance between prediction accuracy and error stability. In the experimental process, based on the Metro Interstate Traffic Volume dataset, this paper completes the whole process of data preprocessing, feature extraction, model training and evaluation, and systematically compares it with many mainstream deep learning models. The experimental results show that the MSTIM model outperforms the comparison models in terms of MAE, MSE and RMSE, verifying its advantages in modeling capability, prediction accuracy and generalization performance.

However, there are still some research limitations in this paper. On the one hand, the current model only models the univariate traffic flow of a single road section, and has not yet fully considered the joint modeling of multiple road sections and the fusion analysis of heterogeneous data from multiple sources (e.g., video streams, real-time event information); on the other hand, the training efficiency and deployment lightness of the model have not yet been solved, and its deployment in the actual ITS system needs to be further optimized.

Therefore, future research will be carried out in the following aspects: first, introducing graph neural network (GNN) or spatial modeling structure to achieve collaborative modeling of traffic flow on multiple road sections; second, integrating multiple sources of information, such as weather, holidays, and emergencies, to improve the model's ability to respond to non-normal hours; third, exploring the model lightweighting strategy and online prediction ability, and enhancing the real-time deployment of the model on the edge equipment Third, to explore the model lightweighting strategy and online prediction capability, and to enhance the real-time deployment capability of the model on edge devices, providing technical support for the construction of a more efficient and intelligent traffic prediction system.
\section*{Acknowledgments}
Thanks for the support provided by the MindSpore Community.

\bibliographystyle{unsrt}  
\bibliography{references}  

\begin{thebibliography}{10}

\bibitem{1}
Jian Wen, Jinhua Zhao, and Patrick Jaillet.
\newblock Rebalancing shared mobility-on-demand systems: A reinforcement learning approach.
\newblock In {\em 2017 IEEE 20th international conference on intelligent transportation systems (ITSC)}, pages 220--225. Ieee, 2017.

\bibitem{2}
Jing Chen, Haocheng Ye, Zhian Ying, Yuntao Sun, and Wenqiang Xu.
\newblock Dynamic trend fusion module for traffic flow prediction.
\newblock {\em arXiv preprint arXiv:2501.10796}, 2025.

\bibitem{3}
Jing Chen, Haocheng Ye, Zhian Ying, Yuntao Sun, and Wenqiang Xu.
\newblock Dynamic trend fusion module for traffic flow prediction.
\newblock {\em arXiv preprint arXiv:2501.10796}, 2025.

\bibitem{4}
Renhe Jiang, Zhaonan Wang, Jiawei Yong, Puneet Jeph, Quanjun Chen, Yasumasa Kobayashi, Xuan Song, Shintaro Fukushima, and Toyotaro Suzumura.
\newblock Spatio-temporal meta-graph learning for traffic forecasting.
\newblock In {\em Proceedings of the AAAI conference on artificial intelligence}, volume~37, pages 8078--8086, 2023.

\bibitem{5}
Yu~Zhao, Pan Deng, Junting Liu, Xiaofeng Jia, and Mulan Wang.
\newblock Causal conditional hidden markov model for multimodal traffic prediction.
\newblock In {\em Proceedings of the AAAI Conference on Artificial Intelligence}, volume~37, pages 4929--4936, 2023.

\bibitem{6}
Shengnan Guo, Youfang Lin, Huaiyu Wan, Xiucheng Li, and Gao Cong.
\newblock Learning dynamics and heterogeneity of spatial-temporal graph data for traffic forecasting.
\newblock {\em IEEE Transactions on Knowledge and Data Engineering}, 34(11):5415--5428, 2021.

\bibitem{7}
Di~Jin, Jiayi Shi, Rui Wang, Yawen Li, Yuxiao Huang, and Yu-Bin Yang.
\newblock Trafformer: Unify time and space in traffic prediction.
\newblock In {\em Proceedings of the AAAI conference on artificial intelligence}, volume~37, pages 8114--8122, 2023.

\bibitem{8}
Mohammad~M Hamed, Hashem~R Al-Masaeid, and Zahi M~Bani Said.
\newblock Short-term prediction of traffic volume in urban arterials.
\newblock {\em Journal of Transportation Engineering}, 121(3):249--254, 1995.

\bibitem{9}
Brian~L Smith and Michael~J Demetsky.
\newblock Traffic flow forecasting: comparison of modeling approaches.
\newblock {\em Journal of transportation engineering}, 123(4):261--266, 1997.

\bibitem{10}
Lun Zhang, Qiuchen Liu, Wenchen Yang, Nai Wei, and Decun Dong.
\newblock An improved k-nearest neighbor model for short-term traffic flow prediction.
\newblock {\em Procedia-Social and Behavioral Sciences}, 96:653--662, 2013.

\bibitem{11}
Jinyoung Ahn, Eunjeong Ko, and Eun~Yi Kim.
\newblock Highway traffic flow prediction using support vector regression and bayesian classifier.
\newblock In {\em 2016 International conference on big data and smart computing (BigComp)}, pages 239--244. IEEE, 2016.

\bibitem{12}
Weilin Ge, Yang Cao, Zhiming Ding, and Limin Guo.
\newblock Forecasting model of traffic flow prediction model based on multi-resolution svr.
\newblock In {\em Proceedings of the 2019 3rd International Conference on Innovation in Artificial Intelligence}, pages 1--5, 2019.

\bibitem{13}
Xiaoyang Wang, Yao Ma, Yiqi Wang, Wei Jin, Xin Wang, Jiliang Tang, Caiyan Jia, and Jian Yu.
\newblock Traffic flow prediction via spatial temporal graph neural network.
\newblock In {\em Proceedings of the web conference 2020}, pages 1082--1092, 2020.

\bibitem{14}
Renteng Yuan, Mohamed Abdel-Aty, and Qiaojun Xiang.
\newblock A study on diversion behavior in weaving segments: Individualized traffic conflict prediction and causal mechanism analysis.
\newblock {\em Accident Analysis \& Prevention}, 205:107681, 2024.

\bibitem{15}
Karthika Balasubramani and Uma~Maheswari Natarajan.
\newblock Improving bus passenger flow prediction using bi-lstm fusion model and smo algorithm.
\newblock {\em Babylonian Journal of Artificial Intelligence}, 2024:73--82, 2024.

\bibitem{16}
Karimeh Ibrahim~Mohammad Ata, Mohd~Khair Hassan, Ayad~Ghany Ismaeel, Syed Abdul~Rahman Al-Haddad, Sameer Alani, et~al.
\newblock A multi-layer cnn-gruskip model based on transformer for spatial- temporal traffic flow prediction.
\newblock {\em Ain Shams Engineering Journal}, 15(12):103045, 2024.

\bibitem{17}
Christofel~Rio Goenawan.
\newblock Astm: Autonomous smart traffic management system using artificial intelligence cnn and lstm.
\newblock {\em arXiv preprint arXiv:2410.10929}, 2024.

\bibitem{18}
Agnimitra Sengupta, Adway Das, and S~Ilgin Guler.
\newblock Hybrid hidden markov lstm for short-term traffic flow prediction.
\newblock {\em arXiv preprint arXiv:2307.04954}, 2023.

\bibitem{19}
J.~Hogue.
\newblock Metro interstate traffic volume [data set].
\newblock \url{https://archive.ics.uci.edu/ml/datasets/Metro+Interstate+Traffic+Volume}, 2019.
\newblock UCI Machine Learning Repository.

\end{thebibliography}

\end{document}